\documentclass[letterpaper]{article} 
\usepackage{aaai24}  
\usepackage{times}  
\usepackage{helvet}  
\usepackage{courier}  
\usepackage[hyphens]{url}  
\usepackage{graphicx} 
\usepackage{natbib}  
\usepackage{caption} 
\usepackage{subfigure}
\usepackage{amssymb}
\usepackage{textcomp}
\usepackage{amsmath}
\usepackage{mathtools}
\usepackage{algorithm}
\usepackage{algorithmic}

\usepackage{newfloat}
\usepackage{listings}
\DeclareCaptionStyle{ruled}{labelfont=normalfont,labelsep=colon,strut=off} 
\floatstyle{ruled}
\newfloat{listing}{tb}{lst}{}
\floatname{listing}{Listing}
\title{TelTrans: Applying Multi-Type Telecom Data to Transportation Evaluation and Prediction via Multifaceted Graph Modeling}
\author{
    ChungYi Lin\textsuperscript{\rm 1,2}, Shen-Lung Tung\textsuperscript{\rm 1}, Hung-Ting Su\textsuperscript{\rm 2}, Winston H. Hsu\textsuperscript{\rm 2,3}
    \\
}
\affiliations{
    \textsuperscript{\rm 1}Internet of Things Laboratory, Chunghwa Telecom Laboratories\\
    \textsuperscript{\rm 2}National Taiwan University\\
    \textsuperscript{\rm 3}Mobile Drive Technology\\

}

\usepackage{bibentry}

\begin{document}

\maketitle

\begin{abstract}
To address the limitations of traffic prediction from location-bound detectors, we present Geographical Cellular Traffic (GCT) flow, a novel data source that leverages the extensive coverage of cellular traffic to capture mobility patterns. Our extensive analysis validates its potential for transportation. Focusing on vehicle-related GCT flow prediction, we propose a graph neural network that integrates multivariate, temporal, and spatial facets for improved accuracy. Experiments reveal our model's superiority over baselines, especially in long-term predictions. We also highlight the potential for GCT flow integration into transportation systems.
\end{abstract}

\section{Introduction}
Accurate traffic prediction can alleviate congestion \cite{lv2021deep} and enhance traffic signal optimization \cite{xie2020urban} for intelligent transportation
systems. However, traditional traffic prediction approaches that rely on dedicated sensors require costly devices and maintenance, limiting application coverage and leading to insufficient information.

\begin{figure}[h]
\centering
\includegraphics[width=0.95\linewidth]{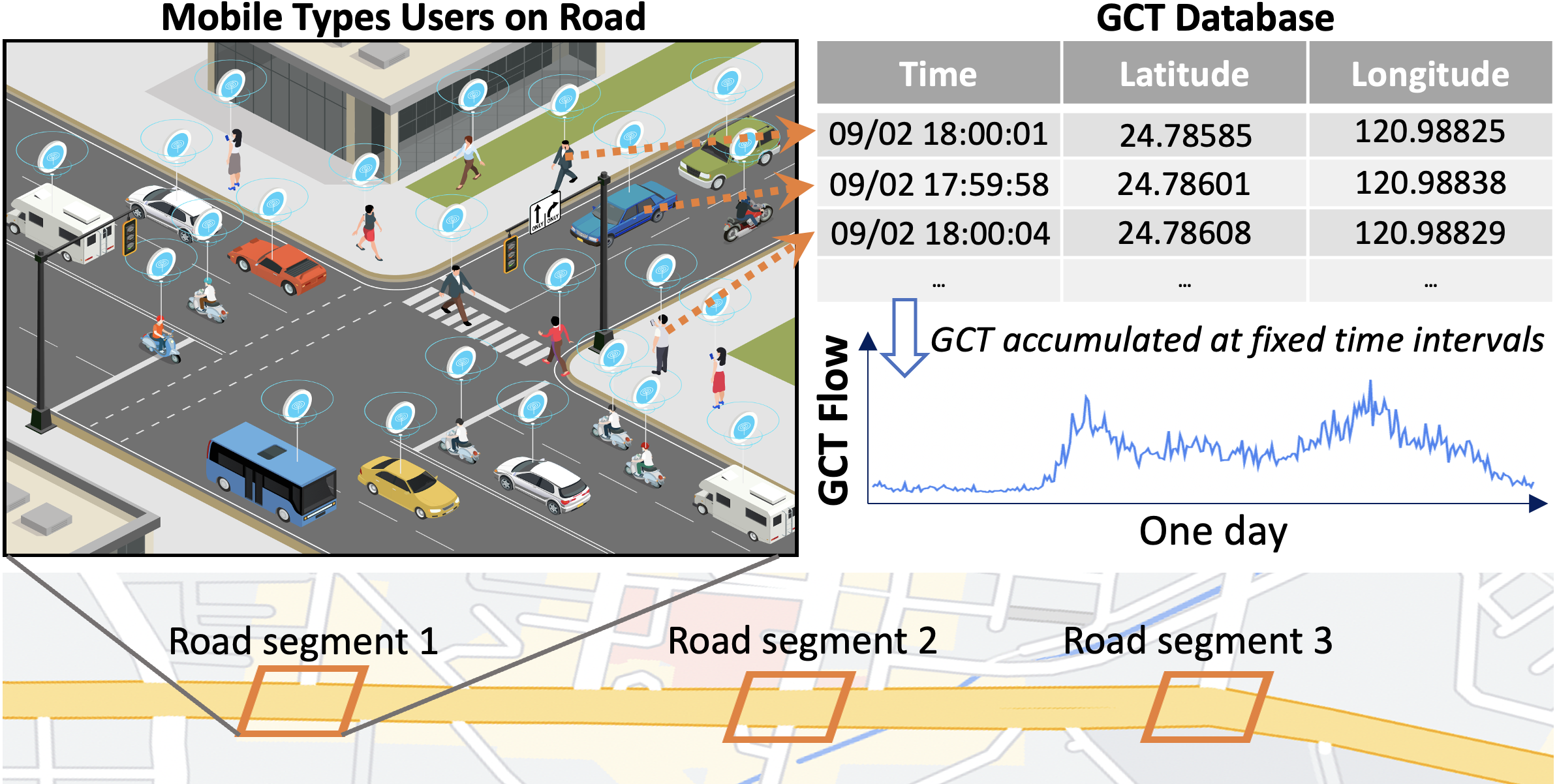}
\caption{An overview of GCT flow. Cellular traffic, generated by mobile users on various road segments, corresponds to entries in the GCT database. The accumulated count of GCT over a given time interval is defined as GCT flow. Over time, the flow series reflects road traffic patterns.}
\label{fig:overview}
\end{figure}

Leveraging the extensive mobile network coverage in Taiwan, we analyzed cellular traffic generated by users' mobile devices \cite{jiang2022cellular,zhao2021spatial}. This geographically-inherent cellular traffic offers a unique insight into transportation dynamics. In partnership with Taiwan's leading telecom company, we present the Geographical Cellular Traffic (\textbf{GCT}). Additionally, we define the \textbf{GCT flow} as the accumulation of GCTs over specific time intervals. To apply GCT flow to transportation, we focus on specific GCT flows collected from various road segments, predicting their trends over time in the relevant areas, as shown in Figure \ref{fig:overview}.

While GCT flow encompasses various user types like vehicles, pedestrians, and stationary users, our primary objective is to evaluate traffic conditions, necessitating a focus on vehicle-related GCT flow. Accordingly, we categorize this flow into three distinct types: \textbf{Vehicle (V-GCT)}, Pedestrian (P-GCT), and Stationary (S-GCT) flows based on telecom company classifications. Each type displays distinct traffic patterns; for instance, commuting routes see V-GCT flow spikes during weekday rush hours, whereas shopping districts might have high P-GCT and S-GCT flows peaks during weekends. As incorporating regional functionalities can improve prediction accuracy \cite{ijcai2022p309}, by exploring the interplays among multi-type GCT flows to uncover implicit functionalities, we can refine our predictive V-GCT.

Predicting V-GCT flow presents \textbf{challenges}. \textit{Firstly}, exploring interactions among multi-type GCT flows adds complexity. \textit{Secondly}, GCT flow fluctuates more than traffic data, suggesting intricate temporal patterns. \textit{Lastly}, as mobile users move across roads, they induce spatial correlations among adjacent roads. Addressing the challenges, we propose a novel model with three \textbf{facets}: \textit{Multivariate}, discerning the interplay among multi-type GCT flows to uncover hidden regional functionality; \textit{Temporal}, separating short-term and long-term dynamics to avoid entangled dependencies \cite{ye2022learning}; and \textit{Spatial}, capturing bidirectional user mobility to understand the spatial dependencies. Experiments confirm the model's efficacy and the importance of integrating multi-type GCT flows. Our key contributions:

$\bullet$ \textbf{Novelty}: We present multi-type GCT flow as a novel method for transportation evaluation, developing a multifaceted graph model for accurate predictions.

$\bullet$ \textbf{Reproducibility}: All datasets and codes are available at: https://github.com/cylin-cmlab/GCT-Prediction.

$\bullet$ \textbf{Predicted Impact}: Our research leads in applying telecom data to traffic management, with V-GCT flow prediction showing great potential for real-world applications.

\clearpage

\section{Multi-type GCT Flows Dataset}
\subsection{Definitions}

\noindent \textbf{Geographical Cellular Traffic (GCT):} Cellular traffic with origin coordinates determined by triangulation \cite{jiang2013review} from the telecom company. Each GCT entry is classified as vehicle, pedestrian, or stationary.

\noindent \textbf{Road Segment:} A 20m x 20m geographical unit for GCT collection, reflecting the typical road widths in Hsinchu, Taiwan, including car lanes, motorcycle lanes, and shoulders. 

\noindent \textbf{Multi-Type GCT Flows:} Quantities of GCT types recorded at fixed intervals (e.g., 5 min) including vehicle (\textbf{V-GCT}), pedestrian (\textbf{P-GCT}), and stationary (\textbf{S-GCT}) flows.
\subsection{Data Collection and Preprocessing}
\textbf{Data sourcing.} 
We extracted full-day GCT data originating within the boundaries of selected road segments, sourced from a major telecom company's database. This database logs over a billion records daily from more than 11 million mobile users, representing approximately 50\% of Taiwan's population. For security purposes, we retained only the crucial data fields without other connection information, as outlined in Table \ref{tab:raw_table}, including the International Mobile Station Equipment Identity (IMEI, a unique identifier for mobile phones), latitude and longitude coordinates, recording time, and the categorized type for each entry.

\begin{table}[h]
  \small
  
  \begin{tabular}{p{1.0cm}p{1.0cm}p{1.2cm}p{1.95cm}p{1.1cm}}
    \hline
    \textbf{IMEI$^{1}$} &  \textbf{Latitude} &  \textbf{Longitude} &  \textbf{Time} &  \textbf{Type$^{2}$}  \\
    \hline
     &   & ... &   &   \\ 
    gnH...mE & 24.78585 & 120.98825 & 09/25 17:59:58 &  vehicle \\ 
    gnB...GI & 24.78601 & 120.98838 & 09/25 18:00:00 & pedestrian \\
    gnK...mU & 24.78608 & 120.98829 & 09/25 18:00:05 & stationary \\
     &   & ... &   &   \\ 
    \hline
  \end{tabular}
  \scriptsize
  $^{1}$IMEI numbers were hashed to protect privacy before processing. \\
  $^{2}$Categorized by the algorithm of the telecom company. 
  \caption{Examples of GCT with essential data fields.}
  \label{tab:raw_table}
\end{table}

\begin{figure}[h]
\centering
\includegraphics[width=0.95\linewidth]{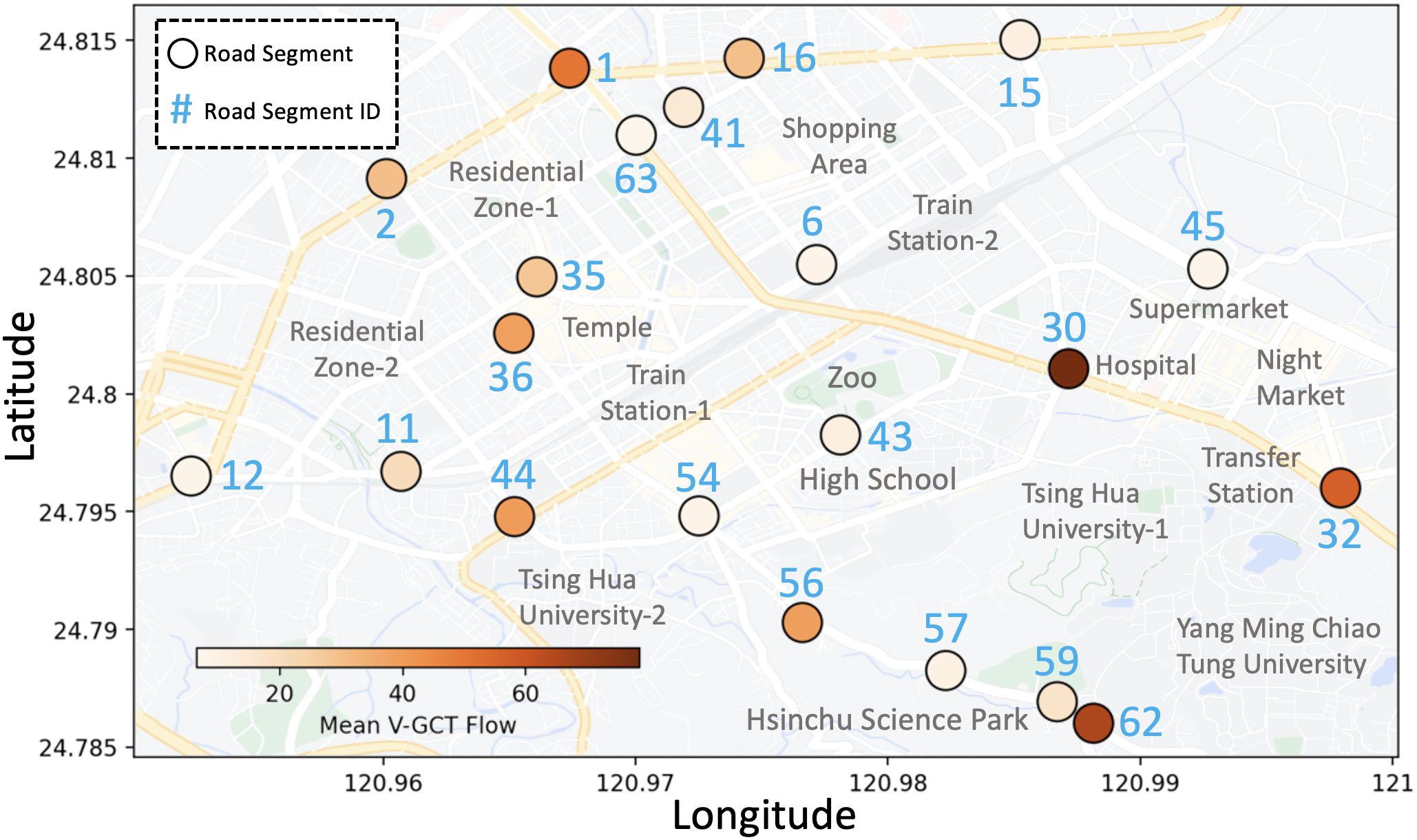}
\caption{Selected road segments in Hsinchu, situated near functional regions or congested routes (e.g., the commuting route from segments 62 to 56), capture representative GCT flow patterns. Segment color indicates the average V-GCT flow from August 28 to September 28, 2022.}
\label{fig_road_segments}
\end{figure}

\noindent \textbf{Road Segments Selection.}
In collaboration with the Hsinchu transportation authority, we selected 21 road segments for our Proof-of-Concept (POC) study, depicted in Figure \ref{fig_road_segments}. The segments were chosen based on insights from local authorities to ensure that the selection captures representative GCT flow patterns or is susceptible to congestion.

\noindent \textbf{Data Cleaning and Processing.}
Duplicate GCT entries with identical timestamps and IMEI numbers were removed to ensure an accurate representation of the GCT flow. Potential issues in cellular network data, such as the 'Ping-Pong' and oscillation effects \cite{zidic2023analyses} (resulting from rapid handovers between cellular base stations) were resolved by the telecom company, enhancing GCT reliability.

\noindent \textbf{Data Privacy Protection.} 
To safeguard user privacy: (1) IMEI numbers in GCT data were hashed, and personal identifiers like user names or addresses were removed, ensuring individuals cannot be distinguished. (2) Data collection was restricted to roads, avoiding business or residential areas to prevent user tracking. (3) The collaborating telecom company adheres to the ISO27001 standard, meaning any data access undergoes a strict approval process, thus preventing unauthorized access.
\begin{table}[ht]
  \small
  \begin{tabular}{p{0.95cm} p{0.63cm} p{0.63cm} p{0.63cm} p{1.4cm} p{1.4cm}}
   \hline
    \textbf{Types}  &  \textbf{Num.} &   \textbf{Avg.} & \textbf{STD}  &  \textbf{Max Avg.} & \textbf{Min Avg.}  \\
   \hline
    V-GCT &  21 & 27.42 & 21.17 & 78.51(ID:30) & 6.49(ID:63) \\
    P-GCT &  21 & 6.86 & 6.38 & 18.03(ID:30) & 1.25(ID:57) \\
    S-GCT &  21 & 9.14 & 7.72 & 23.46(ID:30) & 1.24(ID:57) \\
   \hline
  \end{tabular}
  \caption{Descriptive statistics of multi-type GCT flows.}
  \label{tab:statistics}
\end{table}

\subsection{Data Analysis of GCT flow}
\label{Data_Analysis}
\noindent \textbf{Descriptive Statistics.} Multi-type GCT flow statistics are summarized in Table \ref{tab:statistics}. Road segment 30, situated near the hospital and on main arterial routes, exhibited the highest average for V-(78.51), P-(18.03), and S-(23.46) GCT flows. In contrast, road segment 63, located in a residential zone and only active during rush hours, recorded the lowest average V-GCT at 6.49. Road segment 57, which is part of a commuting route, had the minimum P-(1.25) and S-(1.24) GCT flows. These statistics reveal the variability of multi-type GCT flows across different road segments, reflecting their unique regional functions.

\noindent \textbf{Spatial Coverage.} Our road segment selections in Hsinchu cover pivotal areas including colleges, hospitals, shopping areas, science parks, residential zones, and train stations, capturing representative GCT flows and diverse patterns. Figure \ref{fig_road_segments} shows segments with higher average V-GCT flows, primarily on major arterial routes (e.g., segments 62 to 56 and 30 to 30), reflecting typical urban traffic dynamics with main routes and hubs carrying substantial traffic \cite{peng2016computational,babu2020toward}. The cellular network's wide coverage offers scalability for future road segment selection. 

\noindent \textbf{Temporal Coverage.} Table \ref{tab:statistics} shows that each GCT flow type consists of 8928 samples, gathered by accumulating the respective GCT entries in 5-minute intervals. This results in uniform sample sizes across all GCT flows, providing a consistent depiction of traffic conditions. All type of GCT flows span from August 28 to September 28, 2022, and covers the entire day from 00:00 to 24:00.

\begin{figure}[h]
\centering
\includegraphics[width=0.95\linewidth]{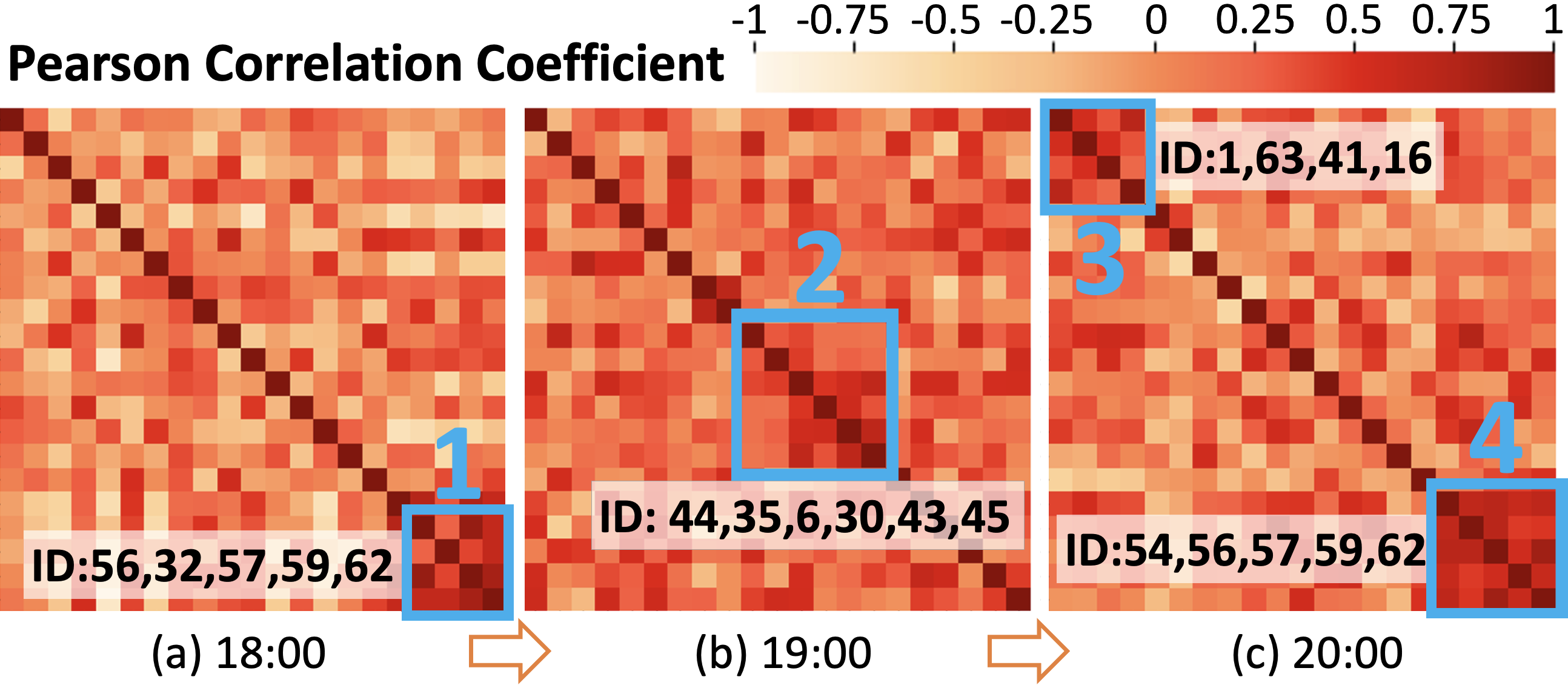}
\caption{Pearson Correlation Coefficients were calculated between pairs of road segments based on 1-hour V-GCT flow at 2022/9/25. As mobility patterns change over time, areas with high coefficients (i.e., blue boxes) shift among different segments, unveiling the evolving spatial correlations.}
\label{fig:pearson}
\end{figure}

\noindent \textbf{Time-Evolving Spatial Correlations.}
Inspired by \cite{zhang2018citywide,wang2021modeling}, we explored the correlation between neighboring segments using the Pearson correlation coefficient \cite{cohen2009pearson} and tracked how these correlations evolved over time. We used the previous 1-hour V-GCT flow of each time slot, to calculate Pearson coefficients among segments, as shown in Figure \ref{fig:pearson}. Neighboring segments typically showed high correlation coefficients due to similar patterns, especially near specific points of interest. 

Figure \ref{fig:pearson}(a) to (c) show evolving spatial correlations, with related segment ranges highlighted in blue boxes. At 18:00, segments near the Science Park (Box 1) had high Pearson coefficients, due to shared commuting patterns as employees left work. By 19:00, segments on commuting routes (Box 2) showed strong correlations, indicating collective movement from workplaces to residential zones. At 20:00, correlations increased near residential zones (Box 3) and the Science Park (Box 4), suggesting people returning home or dining nearby and late-working employees leaving work again, respectively. Overall, this analysis highlights the evolving traffic patterns throughout the periods and the significance of spatial correlation in V-GCT flow.

\subsection{Interactions among GCT Flow’s Types} With analyses of the GCT flows from different perspectives, we aim to discern hidden interactions between user groups.

\begin{figure}
  \centering
  
  \subfigure[Commuting route.]{\label{fig:pattern_62}\includegraphics[width=0.47\linewidth]{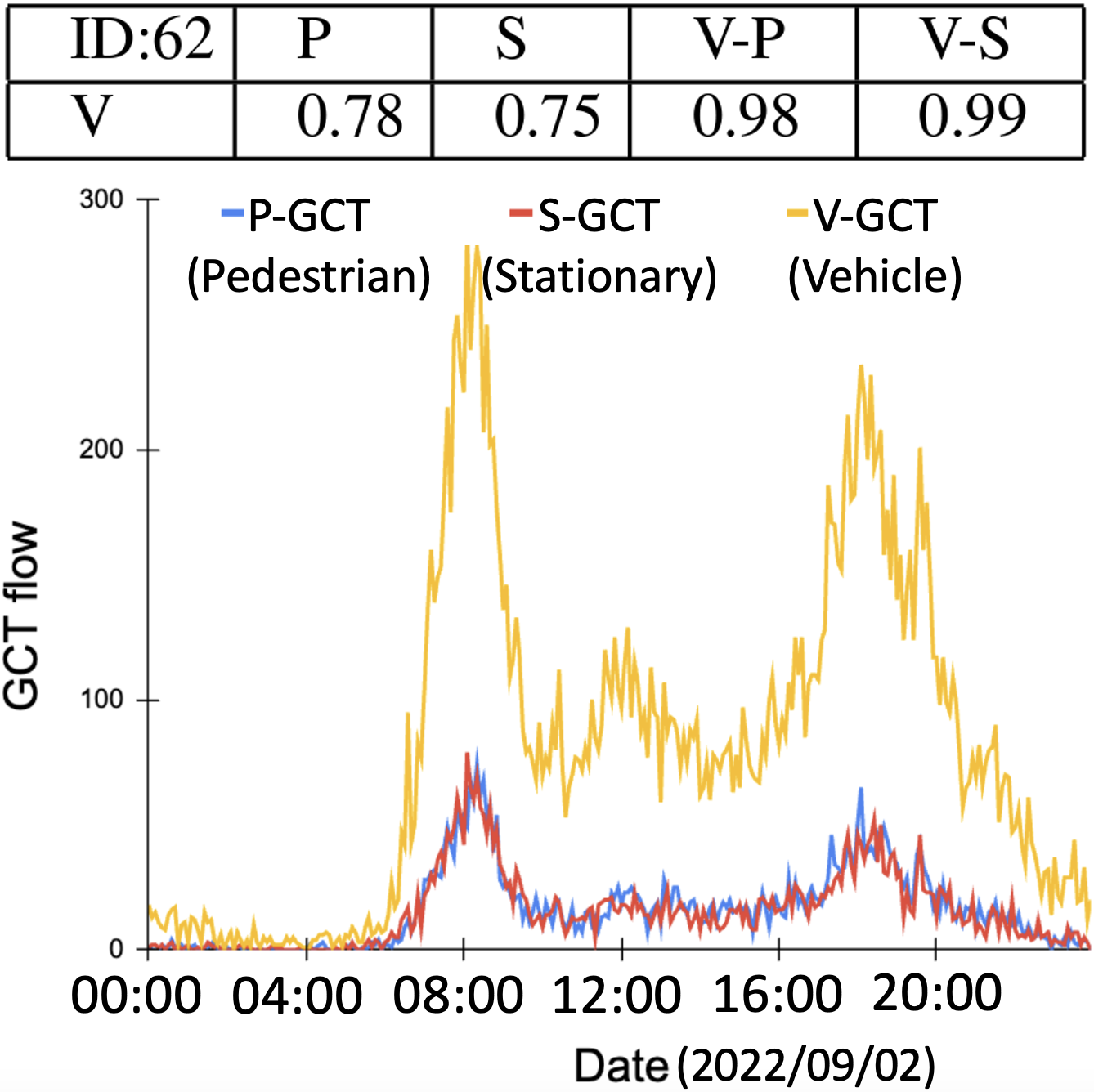}
     
  }
  \subfigure[Residential zone.]{\label{fig:pattern_41}\includegraphics[width=0.47\linewidth]{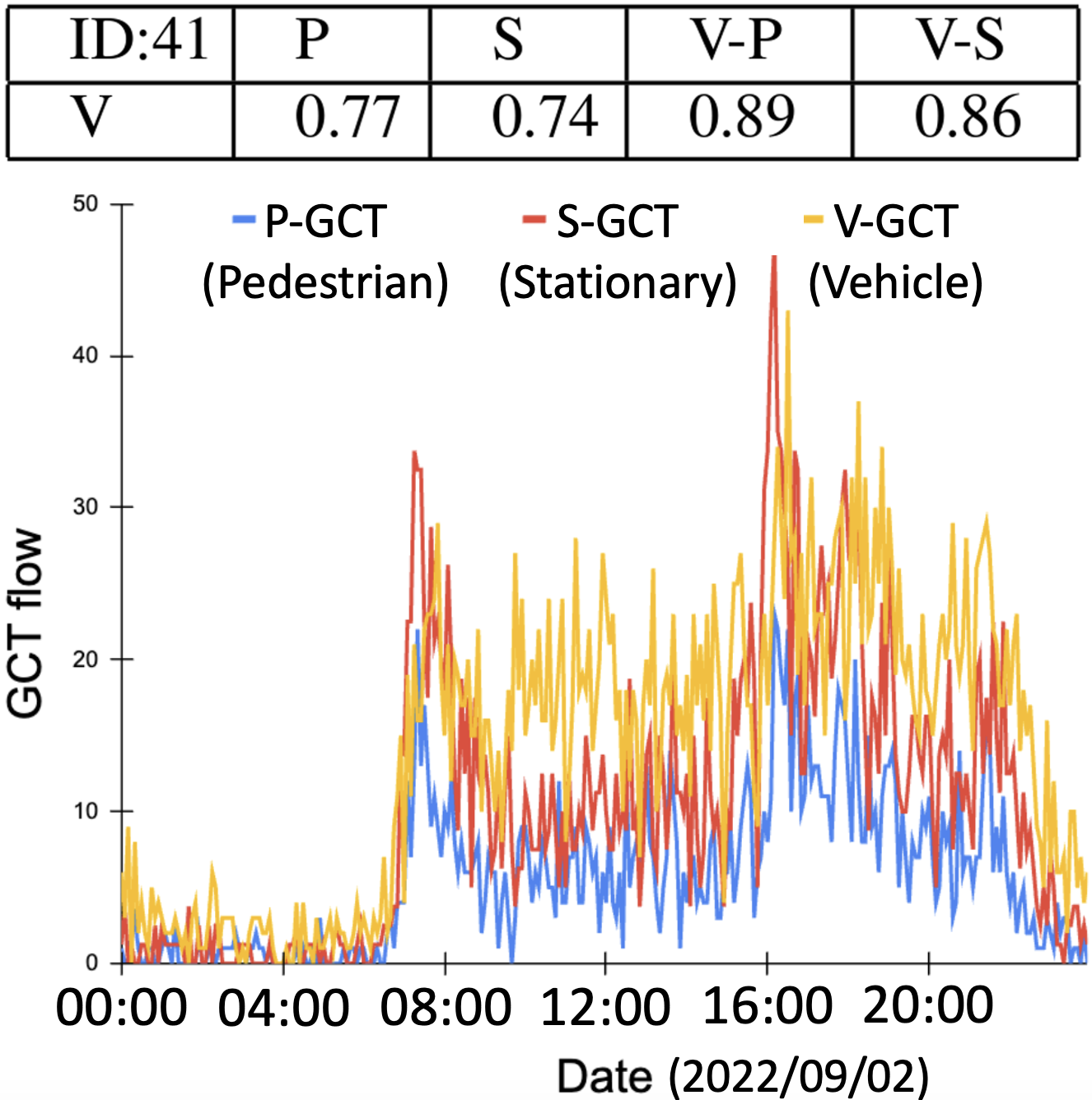}
     
  }
  \caption{Despite distinct GCT flow patterns in commuting routes and residential zones, Pearson correlation coefficients with 1-day data between V-GCT (V) and P-GCT (P) or S-GCT (S) are similar across both segments: 0.78 (P) and 0.75 (S) vs. 0.77 (P) and 0.74 (S). However, subtracted flows (V-P and V-S) yield more noticeable differences: 0.98(V-P) and 0.99(V-S) versus 0.89(V-P) and 0.86(V-S). This suggests that using subtracted flows can reveal more nuanced insights than directly exploring correlations among GCT flows.}
  \label{fig:functionality}
\end{figure}

\noindent \textbf{Exhibiting Regional Pattern.} Figures \ref{fig:pattern_62} show that the commuting route has a dominant V-GCT flow relative to P-GCT and S-GCT flows. In contrast, commercial areas display significant P-GCT and S-GCT flows in Figure \ref{fig:pattern_41}.

\noindent \textbf{Introducing Subtracted Types.}
\label{subtract}
Direct exploring correlations among multivariate data might overlook latent associations \cite{zhao2020multivariate,deng2021graph}. To uncover implicit relationships of user group densities, we introduced subtracted flow types. By subtracting P-GCT and S-GCT from V-GCT, we obtained \textbf{(V-P)} and \textbf{(V-S)} flows, which highlight user group density differences and offer a new perspective of traffic patterns. As shown in Figure \ref{fig:functionality}, in the commuting route (ID:62) of Hsinchu Science Park, where V-GCT flow dominates P-GCT and S-GCT flows, pronounced (V-P) and (V-S) disparities arise.

Using the Pearson correlation coefficient, we analyzed the relationships between V-GCT and P- or S-GCT flows, as well as between V-GCT and its subtracted types (V-P or V-S), as shown in Figure \ref{fig:functionality}. Notably, the difference in Pearson's correlation coefficient between V-GCT and its subtracted flows in different regions (such as commuter routes or residential areas) is more than nine times larger than that observed when directly exploring the differences in correlations between the above multi-type GCT flows. This suggests that examining the association between V-GCT and its subtracted flow reveals the underlying regional attributes.

\subsection{Path Forward for Real-World Deployment}
\begin{figure}[ht]
\centering
\includegraphics[width=0.95\linewidth]{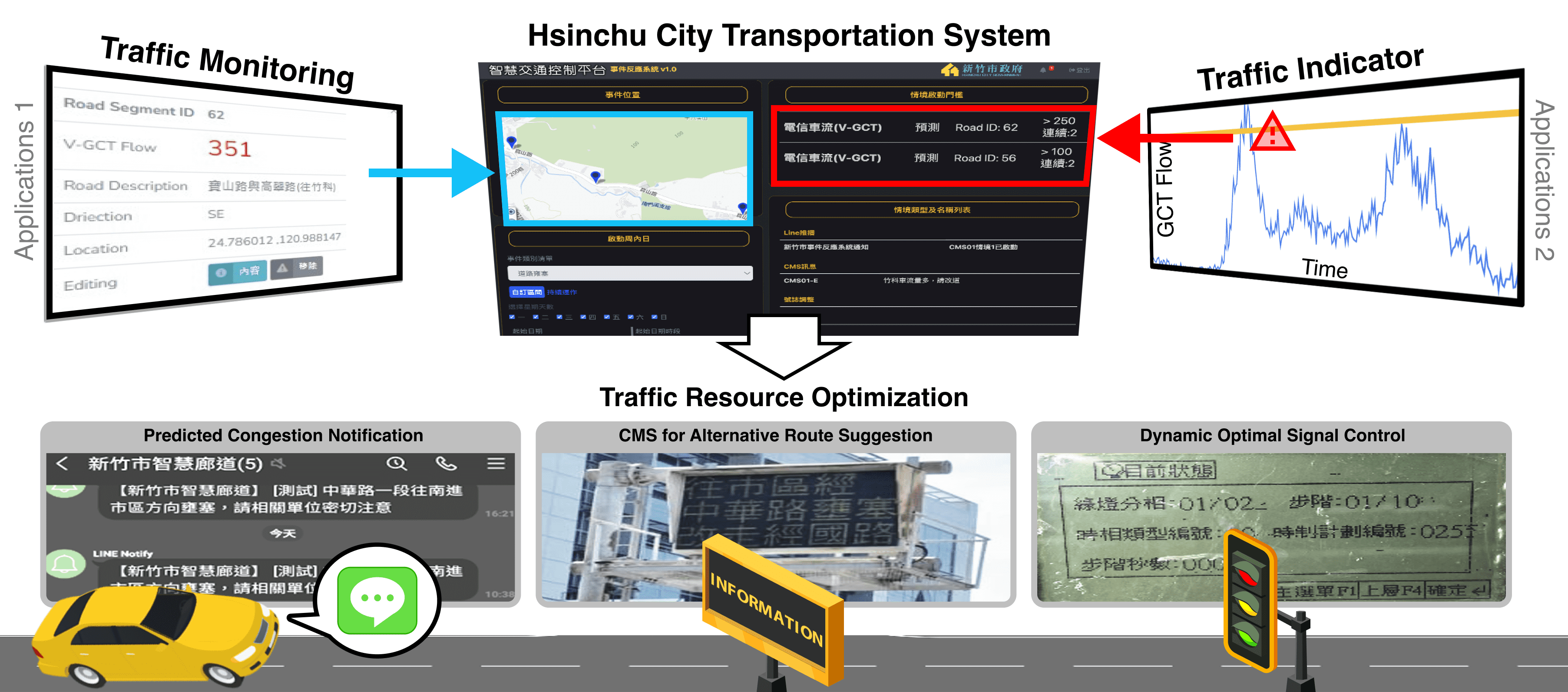}
\caption{System integration of GCT flow enables real-time crowd monitoring and threshold alerts. Exceeding city-defined limits triggers traffic optimization strategies.}
\label{fig:applications}
\end{figure}

Based on the promising potential for applying GCT flow to transportation, we're preparing to integrate its predictive capabilities into the Hsinchu Transportation system. We work with city authorities with two focuses: real-time crowd density monitoring and a threshold-based traffic optimization system, detailed in Figure \ref{fig:applications}. The former will gauge congestion at events such as parades, assisting with resource allocation. The latter, activated when flow surpasses pre-set limits, will trigger measures such as mobile alerts, alternate route suggestions via Changeable Message Signs (CMS), and traffic signal adjustments. These strategies aim to improve traffic insights and efficiently manage congestion.

\clearpage

\section{MultiFaceted Graph Modeling (MFGM)}

\subsection{V-GCT Flow Prediction Task Definition} 
Our objective is to use the historical $T_{in}$ steps of multi-type GCT flow from $N$ road segments, represented by the $\mathcal{X}=\{X_{f^{'}}^{1},...,X_{f^{'}}^{T_{in}}\}$, to forecast V-GCT: $\{{Y}^{T_{in}+1},...,{Y}^{T_{in}+T_{out}}\}$, for $N$ segments in the upcoming $T_{out}$ steps. Here, $f^{'}$ stands for V-GCT, S-GCT, or P-GCT.

\subsection{Overview of the Proposed Model}
\begin{figure}[h]
\centering
\includegraphics[width=0.9\linewidth]{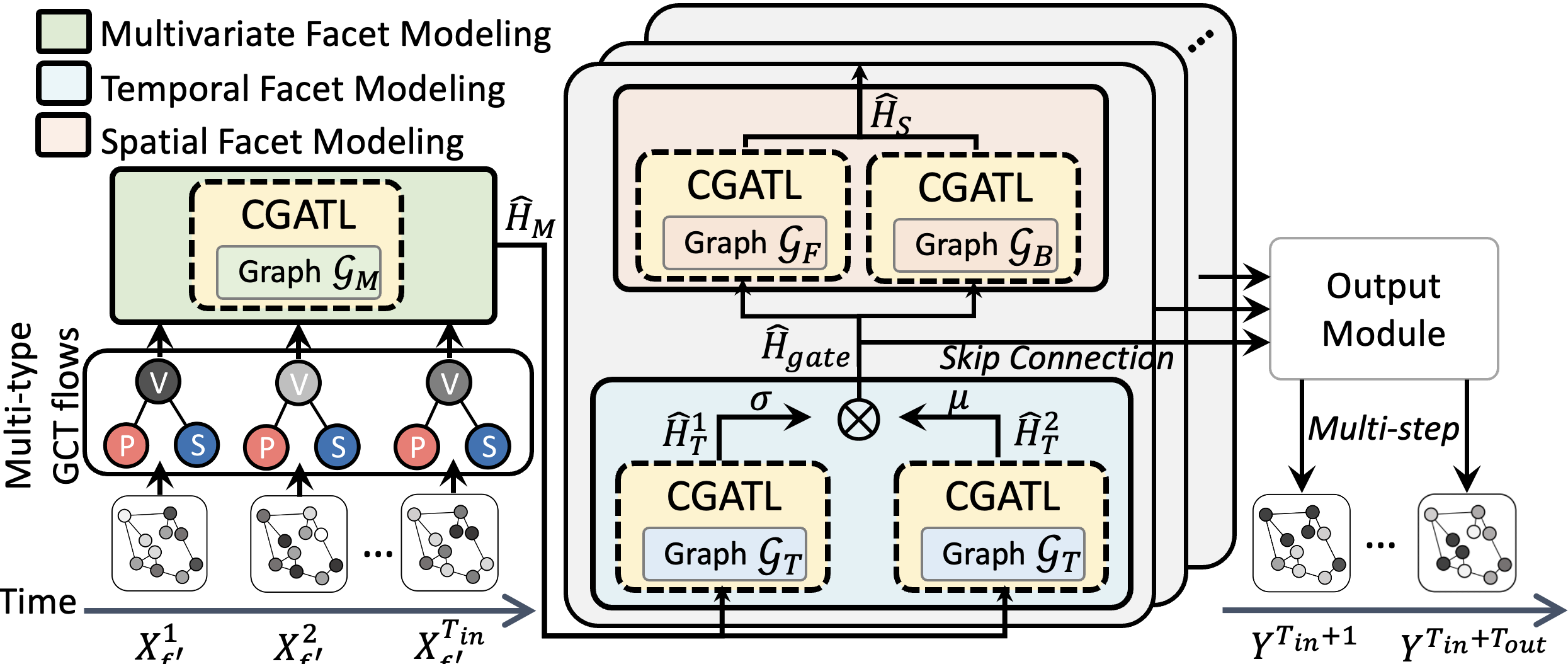}
\caption{Our proposed model consists of three views: Multivariate, Temporal, and Spatial, each exploring correlations from different facets of GCT flow. Each of these modules is based on the Channel-specific Graph Attention Layer (CGATL) to enhance the capability of GAT. The Output Module, connected through skip connections, generates multi-step predictions.}
\label{fig:model}
\end{figure}

Figure \ref{fig:model} shows the proposed model consists of three \textbf{facets}:

\textbf{\textit{Multivariate Facet.}} Capturing interactions among multi-type GCT flows reveals implicit regional functionality.

\textbf{\textit{Temporal Facet.}} Extracts short-term and long-term patterns to discern sudden and regular patterns separately.

\textbf{\textit{Spatial Facet.}} Captures bidirectional spatial dependencies among road segments due to user mobility.

\textbf{\textit{Channel-Specific Graph Attention (CGAT).}} The above facet modeling utilizes a new Graph Attention-based module as its core module, termed Channel-specific Graph Attention (CGAT). CGAT offers flexibility for varying input shapes across facets while retaining efficiency.
\subsection{Channel-Specific Graph Attention Layer (CGATL)}
\noindent \textbf{Preliminaries.}
For \textbf{spatio-temporal input} of $N$ nodes (e.g., road segments) with $D$ observations (e.g., temporal sequences), current spatio-temporal (ST) models \cite{wu2019graph, wu2020connecting, han2021dynamic, ye2022learning} employ a CNN layer with $C$ 1D kernels for feature extraction, producing a \textbf{multi-channel representation}: $H=\{h_{1},h_{2},\dots,h_{N}\} \in \mathbb{R}^{C \times N \times D}$. Each channel, or feature map, captures specific spatio-temporal patterns (e.g., sudden events or periodic rush hours). However, existing Graph Attention Networks (GAT) \cite{velivckovic2018graph,brody2022how} uniformly weight these channels \cite{zhang2020spatial}, risking omission of critical patterns. Thus, we propose an enhanced GAT for handling such representation effectively.

\noindent \textbf{CGAT Mechanism.}
For channel representation (size [$C \times N \times D$]), we employ $C$ independent GATs, to examine distinct correlations among $N$ nodes of each. Considering the $c$-th channel, $H^{c}=\{h^{c}_{1},h^{c}_{2},\dots,h^{c}_{N}\} \in \mathbb{R}^{N \times D}$, the \textit{attention coefficient} in GAT,
\begin{math}
e(h^{c}_{i},h^{c}_{j}),
\end{math}
assesses the dynamic importance of node $j$ to node $i$. Notably, we omit the weight matrix in GAT \cite{velivckovic2018graph,brody2022how} to avoid impacting CNN kernel weights.

Subsequently, we normalize these coefficients across all neighbors of node $i$. This results in the \textit{attention score}, denoted as: $\alpha^{c}_{ij} = softmax(e(h^{c}_{i},h^{c}_{j}))$. We then compute a weighted sum of the features for node $i$ and its neighbors. The outputs from $C$ independent attention mechanisms are then concatenated as follows:
\begin{equation}
\hat{h}_{i}=\|_{c=1}^{C}(\sigma(\sum_{j \in N_{i}}\alpha^{c}_{ij}{h}^{c}_{j})),
\label{eq:channel_concat}
\end{equation}
where $i\in \{1,2,\dots,Z\}$, $N_{i}$ is the neighbors of node $i$, and $\sigma(\cdot)$ is a nonlinear function. We denote the function representing the concatenated $\hat{h}_{i}$ as:
\begin{equation}
CGAT(H,\mathcal{G}) \coloneqq \{\hat{h}_{1},\hat{h}_{2},...,\hat{h}_{N}\},
\label{eq:mgat}
\end{equation}
where $H$ with dimension [$C \times N \times D$] and $\mathcal{G}$ is a graph structure indicating the connections among nodes in $H$.

\noindent \textbf{CGAT Layer (CGATL) for Efficiency Management.}
For computational efficiency when using CGATL on high channel-count representations, like the 32 channels in \cite{wu2020connecting}, we adopt 1$\times$1 convolution layers for Encoder and Decoder, inspired by the sandwich structure \cite{yu2018spatio}. The Encoder reduces the channel count from $C$ to $C^{'}$ ($C^{'}$ $<$ $C$), processes through CGAT, and then the Decoder restores the count to $C$. Residual connections between the Encoder and Decoder ensure training stability. The \textbf{CGATL Layer (CGATL)} is given by:
\begin{align}
CGATL(H,\mathcal{G}) &\coloneqq \nonumber \\
Decoder(&CGAT(Encoder(H),\mathcal{G})) + H.
\label{eq:all_channel_aggregation_symbol}
\end{align}
We then present how the CGATL is applied to various facets.

\subsection{Multivariate Facet Modeling}
\label{multivariate_view_modeling}
\noindent \textbf{Insights.}
Modeling directly on relationships in multivariate data may not adequately highlight correlations, as depicted in Figure \ref{fig:functionality}. By modeling with subtracted type flows, we emphasize differences between V-GCT and P-GCT, revealing regional attributes. This method enhances prediction accuracy, as confirmed by our experiments.

\noindent \textbf{Notations:}

$\bullet$ $H_{vgct}$ and $H_{f}$: The multi-channel representations of V-GCT flow and either P-GCT or S-GCT flow, respectively. $f$ can either be P-GCT or S-GCT flow.

$\bullet$ $\Delta_{f}$: Represents the subtraction-type flow, which is computed as $\Delta_{f} = H_{vgct} - H_{f}$.

$\bullet$ $\mathcal{G}_{M}$: Represents a complete graph that reveals the connections among various GCT flow types.

\noindent \textbf{Implementation.}
For each time step $t$, we concatenate $H_{vgct}^{t} \in \mathbb{R}^{C \times N \times 1} $ with $\Delta_{pgct}^{t} \in \mathbb{R}^{C \times N \times 1}$ and $\Delta_{sgct}^{t} \in \mathbb{R}^{C \times N \times 1}$ along the third dimension. This produces a representation of size [$C \times N \times 3 $], which we then reshape into [$C \times 3 \times N $], denoted as $H_{\Delta}^{t}$, to align with the CGATL procedure. After CGATL, we extract only the 'enhanced' $H_{vgct}^{t}$:
\begin{equation}
\hat{H}_{vgct}^{t} = Extract( CGATL(H_{\Delta}^{t},\mathcal{G}_{M})),
\label{eq:multivariate_modeling}
\end{equation}
where 'Extract' function retrieves the enhanced representation of $H_{vgct}^{t}$, the first element of $H_{\Delta}^{t}$ from CGATL's output. 

Then, we concatenate the output of all time step in Equation \ref{eq:multivariate_modeling} along the second dimension of [$C \times T \times N$] as follows:
\begin{equation}
\hat{H}_{M}=\|_{t=1}^{T}(\hat{H}_{vgct}^{t}).
\label{eq:temporal_concat}
\end{equation}
$\hat{H}_{M} \in \mathbb{R}^{C \times T \times N}$ is output of Multivariate facet Modeling. We integrate the correlations among multi-type GCT flows over time steps while preserving their chronological order.

\subsection{Temporal Facet Modeling}
\noindent \textbf{Insights.}
Complex time series can easily entangle temporal dependencies, making it challenging to identify valuable signals \cite{xu2017real, ye2022learning}. Given the heightened fluctuations in V-GCT flow compared to traditional traffic data like speed, capturing both short-term and long-term patterns is vital. Thus, our model emphasizes understanding these temporal dynamics.

\noindent \textbf{Notations:}

$\bullet$ \textbf{$H$:} Multi-channel representation reshaped to [$C \times T \times S$], regarded as $T$ time nodes with $S$ road segment states.

$\bullet$ \textbf{$\mathcal{G}_{T}$:} The complete graph composed of the $T$ nodes.

\noindent \textbf{Implementation.}
Following \cite{wu2020connecting, ye2022learning}, we use two CNNs with kernel sizes ($2 \times 1$) and ($5 \times 1$) to extract short- and long-term temporal patterns from multi-channel representation $H$, as $H_{2}$ and $H_{5}$, respectively. $H_{2}$ captures local dependencies, while $H_{5}$ reflects longer trends.

After extracting these different scale representations $H_{2}$ and $H_{5}$, we apply CGATL to process them and concatenate:
\begin{equation}
\hat{H}_{T} = Concat([CGATL(H_{2},\mathcal{G}_{T}),CGATL(H_{5},\mathcal{G}_{T})]),
\label{eq:temporal_concat2}
\end{equation}
where 'Concat' fuction merges two outputs along the second dimension of [$C \times T \times N$]. Following \cite{wu2020connecting}, we truncate longer temporal outputs to match shorter ones. Specifically, we reshape $H_{2}$ from [$C \times (T-1) \times S$] to [$C \times (T-4) \times S$] by removing its first three elements, aligning it with $H_{5}$.

To manage the ratio of information passed to the next module, we implement the \textit{Gating Mechanism} \cite{wu2019graph, ye2022learning} by employing dual CGATL to fuse two outputs as follows:
\begin{equation}
\hat{H}_{gate} = \sigma(\hat{H}^{1}_{T}) \odot \mu(\hat{H}^{2}_{T}),
\end{equation}
where $\hat{H}^{1}_{T}$ and $\hat{H}^{2}_{T}$ are generated from Equation \ref{eq:temporal_concat2} in dual CGATL, respectively. $\mu$ denotes the tangent hyperbolic function, and $\odot$ represents the Hadamard product. $\hat{H}_{gate}$ is the fusion output from the Temporal Facet Modeling, which is then fed forward to the next spatial modeling.

\subsection{Spatial Facet Modeling}
\noindent \textbf{Insights for Modeling.}
Our analysis in Figure \ref{fig:pearson} reveals spatial correlations between road segments, where the GCT flow of an upstream segment influences the downstream GCT flow. Additionally, users' mobility may cause them to move back and forth due to their activity behaviors, such as daily commuting patterns to and from work or adjusting their routes when faced with unexpected traffic situations. Hence, considering not only the spatial correlation but also accounting for bidirectional variations among road segments can lead to a comprehensive understanding of user mobility.

\noindent \textbf{Notations:}

$\bullet$ \textbf{$H$:} Multi-channel representation reshaped to [$C \times S \times T$], regarded as $S$ road segment with $T$ observations.

$\bullet$ $A$: The adjacency matrix formed by the $S$ segments.

$\bullet$ $\mathcal{G}_{F}$ and $\mathcal{G}_{B}$: Forward and backward graph structure, respectively, created by processing $A$ with the thresholded Gaussian kernel \cite{shuman2013emerging}. They are computed as $A/rowsum(A)$ and $A^{T}/rowsum(A^{T})$, respectively, for direction-specific spatial modeling.

\noindent \textbf{Implementation.}
To accommodate bidirectional user mobility across road segments among V-GCT flows, we utilize two distinct CGATLs, each used to explore propagation in one direction. We assign specific directional graphs, $\mathcal{G}{F}$ and $\mathcal{G}{B}$, to each CGATL. By feeding the two transition matrices into their corresponding CGATs, we combine them as:
\begin{equation}
\hat{H}_{S} = CGATL(H,\mathcal{G}_{F}) + CGATL(H,\mathcal{G}_{B})
\end{equation}
where $\hat{H}_{S}$ is the fusion output from the Spatial Facet Modeling, which is then fed forward to the next layer.

\subsection{Overall Structure}
Our model begins with a 32-channel CNN layer to encode all GCT flow types. It comprises three layers, each integrating Spatial and Temporal Facet Modeling. Post every Temporal Facet Modeling, we use 64-channel CNN layer as skip connections that lead into the Output Layer. This layer processes the merged representations to derive multi-step outputs. 

\begin{table*}[ht]
\centering
\small

\begin{tabular}
{@{\extracolsep{\fill}}  p{0.92cm}  
p{0.84cm}  p{0.84cm} p{1.14cm}  
p{0.84cm}  p{0.84cm} p{1.14cm}
p{0.84cm}  p{0.84cm} p{1.14cm}
p{0.84cm}  p{0.84cm} p{1.14cm}
}
\hline
&  & \textbf{15 min.} &  &   &   \textbf{30 min.}  &   &      &    \textbf{45 min.}  &   &      &   \textbf{60 min.} & \\
 \textbf{Baselines} & MAE & RMSE & MAPE & MAE & RMSE & MAPE & MAE & RMSE & MAPE & MAE & RMSE & MAPE \\
\hline
TCN$^{0}$ 
& 5.55$\pm$.02	& 8.82$\pm$.04 & 34.5\%$\pm$.6 
& 5.74$\pm$.02 & 9.38$\pm$.06 & 36.9\%$\pm$.9
& 6.01$\pm$.02 & 1.05$\pm$.11 & 38.2\%$\pm$.5
& 7.18$\pm$.05 & 12.22$\pm$.1 & 39.7\%$\pm$1.4 \\

GWNet$^{2}$ & 5.46$\pm$.01	& 8.72$\pm$.04 & 32.5\%$\pm$1.3 
& 5.62$\pm$.03 & 9.08$\pm$.11 & 32.9\%$\pm$1.3
& 5.70$\pm$.02  & 9.44$\pm$.03 & 33.8\%$\pm$1.2
& 6.08$\pm$.06& 10.31$\pm$.2 & 34.6\%$\pm$1.6 \\

Gman$^{1}$ & 5.37$\pm$.01	& 8.61$\pm$.05& 32.6\%$\pm$2
& 5.51$\pm$.04 & 8.99$\pm$.14 & 32.8\%$\pm$1.2 
& 5.61$\pm$.02 & 9.20$\pm$.11 & 33.1\%$\pm$.7 
& 5.77$\pm$.02& 9.68$\pm$.11 & 34.4\%$\pm$1.0\\

MTGNN$^{2}$ & 
5.29$\pm$.02 	& 8.52$\pm$.03 & 32.2\%$\pm$1.4 
& 5.45$\pm$.01  & 8.86$\pm$.01 & 32.7\%$\pm$1.6 
& 5.60$\pm$.04  & 9.27$\pm$.03 & 32.9\%$\pm$1.7
& 5.74$\pm$.03   & 9.66$\pm$.13  & 35.3\%$\pm$1.9 \\

MPNet$^{1}$ & 
5.30$\pm$.01 	& 8.46$\pm$.04 & 32.9\%$\pm$2.0 
& 5.44$\pm$.03  & 8.84$\pm$.09 & 34.7\%$\pm$1.6 
& 5.55$\pm$.01  & 9.16$\pm$.02 & 33.1\%$\pm$1.7
& 5.73$\pm$.03   & 9.68$\pm$.06 & 34.8\%$\pm$2.1 \\

DMGCN$^{2}$ 
& 5.28$\pm$.04 & 8.48$\pm$.11 & 31.8\%$\pm$1.7
& 5.46$\pm$.01 & 8.81$\pm$.03 & 33.6\%$\pm$1.9
& 5.68$\pm$.03 & 9.21$\pm$.02 & 33.9\%$\pm$.9
& 5.82$\pm$.02 & 9.56$\pm$.08 & 34.6\%$\pm$1.1 \\

ESG$^{2}$ 
& 5.26$\pm$.03 & 8.43$\pm$.03 & 31.1\%$\pm$1.1
& 5.40$\pm$.02 & 8.76$\pm$.06 & 31.8\%$\pm$1.3
& 5.54$\pm$.03 & 9.16$\pm$.03 & 32.1\%$\pm$1.3
& 5.65$\pm$.02 & 9.46$\pm$.09 & 33.2\%$\pm$1.1 \\

\textbf{MFGM$^{1}$}
& \textbf{5.23$\pm$.01} & \textbf{8.27$\pm$.06} & \textbf{29.8\%$\pm$.7}
& \textbf{5.33$\pm$.02} & \textbf{8.54$\pm$.06} & \textbf{30.5\%$\pm$.8}
& \textbf{5.43$\pm$.05} & \textbf{8.94$\pm$.07} & \textbf{30.9\%$\pm$.6}
& \textbf{5.51$\pm$.03} & \textbf{9.10$\pm$.07} & \textbf{31.3\%$\pm$.7} \\

\hline
\end{tabular}
\\
$^{0}$ denotes the \textbf{TCN-based} methods, $^{1}$ denotes the \textbf{Attention-based} methods, and $^{2}$ denotes the \textbf{GCN-based} methods.

\caption{Performance comparisons from short-term to long-term V-GCT predictions}
\label{table:v_gct_prediction}
\end{table*}
\section{Experiments}
\noindent \textbf{Experimental Setups.}
For the CGATL, channel numbers are set at $C=32$ and $C'=8$. We utilize the Adam optimizer with a learning rate of 0.0005 and weight decay of 0.0001. Following \cite{li2018diffusion}, we split all temporal sequences into 70\% training, 20\% validation, and 10\% testing. Each 24-step sample uses the first 12 steps ($T_{in}$) as historical input, and the following 12 steps ($T_{out}$). More details are at: https://github.com/cylin-cmlab/GCT-Prediction.

\noindent \textbf{Metrics.}
We use Mean Absolute Error (MAE), Root Mean Squared Error (RMSE), and Mean Absolute Percentage Error (MAPE) as evaluation metrics. 

\noindent \textbf{Baselines.} We have chosen representative spatio-temporal baselines for this new task: \textit{Temporal Convolution (TCN)} \cite{yu2015multi}: The convolution method captures dependencies at different time scales. \textit{GWNet} \cite{wu2019graph}: A graph-based WaveNet model equipped with a spatial diffusion mechanism. \textit{MTGNN} \cite{wu2020connecting}: A graph-based convolutional model that dynamically learns graph structures. \textit{Gman} \cite{zheng2020gman}: A graph multi-attention model to capture the impact of the spatio-temporal factors. \textit{MPNet} \cite{lin2021multivariate}: A hybrid-GAT model with propagation attention.
\textit{DMGCN} \cite{han2021dynamic}: A graph-based model that leverages time-specific spatial dependencies. \textit{ESG'22} \cite{ye2022learning}: A graph-based model employing evolutionary and multi-scale graph structures. 

\subsection{V-GCT Prediction Evaluation}
Table \ref{table:v_gct_prediction} presents short- to long-term predictions, categorizing baselines by implementation. Our MFGM consistently outperforms various baselines. In short-term predictions (15-30 mins), recent patterns often prevail, creating similar performances among baselines, yet MFGM still exceeds them. In long-term predictions (45-60 mins), as temporal dependencies and patterns wane \cite{ye2022learning}, MFGM extracts both short- and long-term spatio-temporal dependencies. The former aids near-term predictions, the latter supports distant forecasting, and their integration boosts MFGM's long-term prediction. Furthermore, by leveraging interactions between V-GCT and its subtracted types, our model enhances insights into regional functionality and crowd activity. As predictions extend, MFGM's superiority is emphasized, underlining its real-world applicability.

\subsection{Ablation Study of MFGM}
We evaluate the contributions of MFGM's components by comparing the full model to three ablated versions: without Multivariate Facet Modeling (w/o M), without Spatial Facet Modeling (w/o S), and without Temporal Facet Modeling (w/o T). Table \ref{table:ablation} presents average results for prediction steps 1 (5 min.) to 12 (60 min.), ordered by impact:

\noindent \textbf{Impact of w/o S.} Omitting spatial modeling has the most significant detrimental effect on performance, highlighting the need to capture spatial dependencies between V-GCT flows on road segments.

\noindent \textbf{Impact of w/o M.} Without exploring of hidden correlations between multi-type GCT flows results in the second-worst performance, emphasizing the importance of investigating implicit relationships to reveal regional attributes.

\noindent \textbf{Impact of w/o T.} The less significant performance decline compared to w/o S and w/o M suggests that in predicting V-GCT flow, spatial information and regional functionality are more crucial. However, the performance drop in w/o T still indicates the importance of temporal modeling.

\noindent Overall, the full MFGM consistently surpasses its ablated versions, underlining the essence of each facet modeling.
\begin{table}[h]
  \small
  \centering
  
  \begin{tabular}{c|ccc}
    \hline
    Baselines & MAE & RMSE & MAPE  \\
    \hline
    w/o S &  {5.46}$\pm$0.06	& {8.82}$\pm$0.06 & {32.1\%}$\pm$0.4 \\
    w/o M &  {5.44}$\pm$0.03	& {8.73}$\pm$0.04 & {31.9\%}$\pm$0.5 \\
    w/o T &  {5.40}$\pm$0.03	& {8.71}$\pm$0.06 & {31.5\%}$\pm$0.6 \\
    Full &  \textbf{5.34}$\pm$0.05	& \textbf{8.61}$\pm$0.03 & \textbf{30.9\%}$\pm$0.5 \\
    \hline
  \end{tabular}
  \caption{Ablation study for overall performance.}
  \label{table:ablation}
\end{table}
\subsection{Sensitivity Analysis of Multi-Type GCT Flows}
\begin{figure}[h]
\centering
\includegraphics[width=1\linewidth]{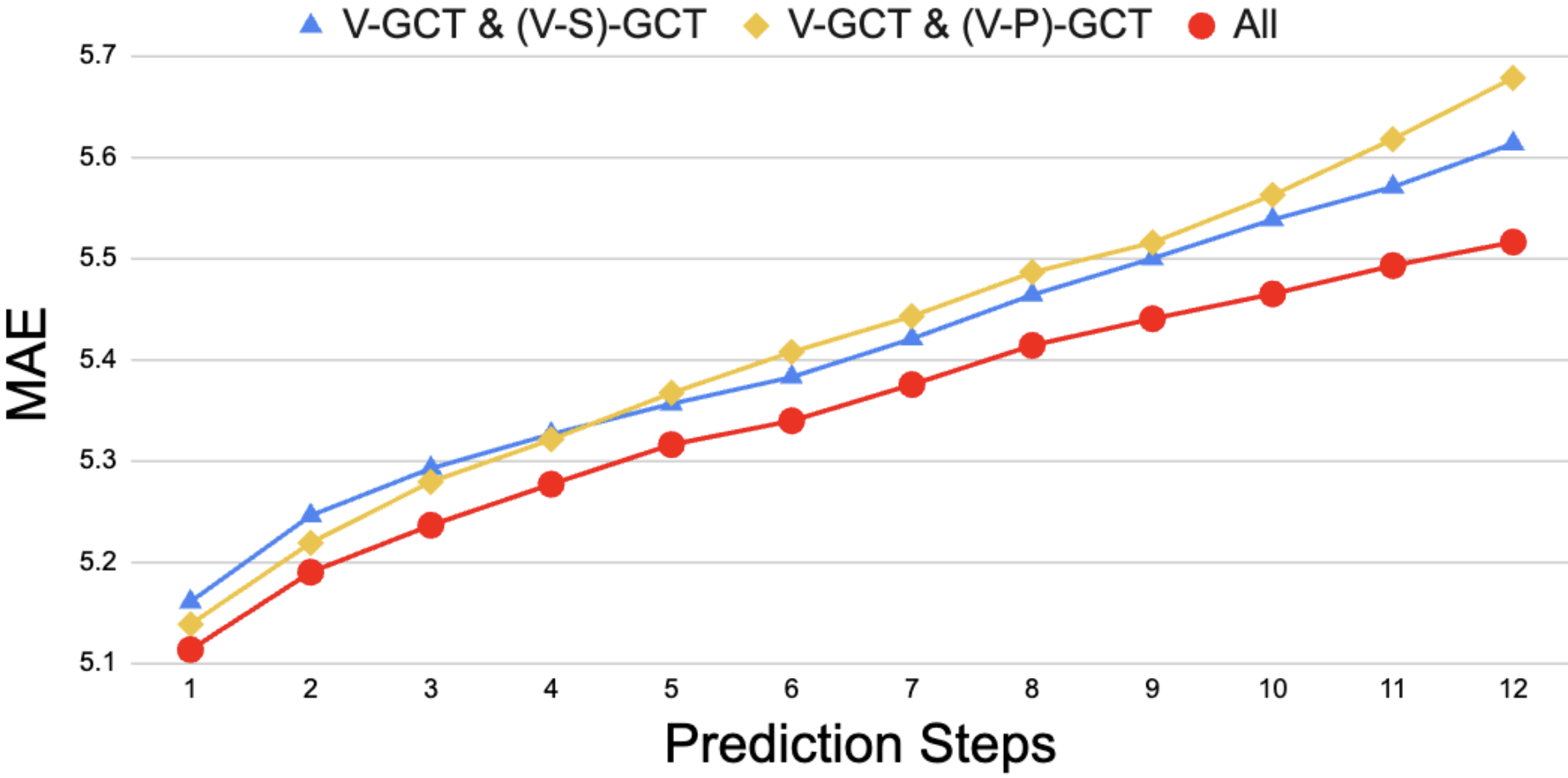}
\caption{The pairing of V-GCT flow with all subtraction-type GCT flows consistently results in the lowest errors.}
\label{fig:sensitivity}
\end{figure}

Figure \ref{fig:sensitivity} shows the effect of different combinations on prediction performance over various steps. In steps 1-4, the model with V-GCT and (V-P)-GCT performs better than the one with V-GCT and (V-S)-GCT. After step 4, the latter model performs better, suggesting varying importance of relative differences between vehicle and pedestrian or stationary GCT flows depending on prediction horizon. Moreover, the model with V-GCT and all subtraction-type GCT flows consistently achieves the top performance across all steps, highlighting the benefits of exploring relative differences among GCT flows.

\section{Conclusion}
We presented multi-type GCT flows as a novel data source for transportation and proposed MFGM to predict V-GCT flows with improved accuracy by integrating multi-type GCT flows. MFGM outperforms baselines, excelling in long-term predictions, and the ablation study confirms the importance of spatial and regional attributes. We also revealed V-GCT integration into transportation systems, presenting new applications for telecom data in transportation.

\clearpage

\section{Acknowledgments}
This work was supported in part by National Science and Technology Council, Taiwan, under Grant NSTC 111-2634-F-002-022 and by Qualcomm through a Taiwan University Research Collaboration Project.

\bibliography{aaai24}

\end{document}